\documentclass[10pt,twocolumn]{article}

\usepackage[T1]{fontenc}
\usepackage[utf8]{inputenc}
\usepackage[protrusion=true,expansion=false]{microtype}
\usepackage{graphicx}
\usepackage{amsmath,amssymb,amsfonts}
\usepackage{booktabs}
\usepackage{multirow}
\usepackage{xcolor}
\usepackage[font=small,labelfont=bf]{caption}
\usepackage{subcaption}
\usepackage{float}
\usepackage{algorithm}
\usepackage{algpseudocode}       
\usepackage{hyperref}
\usepackage{url}
\usepackage[numbers,sort&compress]{natbib}

\usepackage[top=1in,bottom=1in,left=0.75in,right=0.75in,
            columnsep=0.25in]{geometry}

\definecolor{encoder}{RGB}{255,224,178}
\definecolor{graphsage}{RGB}{225,190,231}
\definecolor{decoder}{RGB}{255,249,196}
\definecolor{output}{RGB}{178,223,219}
\definecolor{myblue}{RGB}{0,70,140}

\hypersetup{
  colorlinks=true,
  linkcolor=myblue,
  citecolor=myblue,
  urlcolor=myblue,
  pdftitle={Efficient Hybrid CNN-GNN Architecture for Monocular Depth Estimation},
  pdfauthor={Ishan Narayan}
}

\title{\Large\textbf{Efficient Hybrid CNN-GNN Architecture\\
       for Monocular Depth Estimation}}

\author{
  Ishan Narayan\\
  \small IMCS Lab, CSIR-CSIO\\
  \small Chandigarh 160030, India\\
  \small \texttt{ishan.csio21a@acsir.res.in}
}

\date{}

\usepackage{titlesec}
\titleformat{\section}{\large\bfseries}{\thesection.}{0.5em}{}
\titleformat{\subsection}{\normalsize\bfseries}{\thesubsection.}{0.5em}{}
\titleformat{\subsubsection}{\normalsize\itshape}{\thesubsubsection.}{0.5em}{}

\begin{document}

\maketitle
\thispagestyle{empty}

\begin{abstract}
We present \textbf{GraphDepth}, a monocular depth estimation architecture that
synergistically integrates Graph Neural Networks (GNNs) within a
convolutional encoder-decoder framework.
Our approach embeds efficient GraphSAGE layers at multiple scales of a
ResNet-101 U-Net backbone, enabling explicit modeling of long-range spatial
relationships that lie beyond the receptive field of local convolutions.
Key technical contributions include:
(1) batch-parallelized graph construction with configurable $k$-NN and
    grid-based adjacency for scalable training;
(2) multi-scale GraphSAGE integration at bottleneck and decoder stages
    ($1/32$, $1/16$, $1/8$ resolution) to propagate global context
    throughout the feature hierarchy;
(3) channel-attention gated skip connections that adaptively weight encoder
    features before fusion; and
(4) heteroscedastic uncertainty estimation via a dedicated aleatoric
    uncertainty head, enabling confidence-aware loss weighting during
    optimization.
Unlike transformer-based hybrids, which suffer from quadratic complexity in
sequence length, GraphDepth scales linearly with spatial resolution while
achieving comparable global receptive fields through iterative message
passing.
Experiments on NYU Depth V2, WHU Aerial, ETH3D, and Mid-Air benchmarks
demonstrate competitive accuracy---within 4.6\% of state-of-the-art
transformers on indoor scenes---with substantially lower computational cost
(25 FPS vs.\ 9 FPS, 3.8 GB vs.\ 8.8 GB VRAM).
GraphDepth achieves the best reported result on WHU Aerial (RMSE 8.24 m)
and exhibits superior zero-shot cross-domain transfer to the Mid-Air
synthetic aerial dataset, validating the generalization power of explicit
relational reasoning for depth estimation.
\end{abstract}


\vspace{4pt}
\hrule
\vspace{4pt}

\section{Introduction}
\label{sec:intro}

Monocular depth estimation (MDE) is a fundamental computer vision task
with critical applications in autonomous driving, augmented reality, and
robotic navigation~\cite{eigen2014depth}.
Despite remarkable progress, the field still grapples with a core tension:
\emph{local feature extraction} versus \emph{global context modeling}.

\textbf{CNN limitations.}
Convolutional Neural Networks (CNNs) have long dominated MDE owing to their
inductive biases for spatial locality and computational efficiency.
However, their bounded receptive fields constrain the capture of long-range
geometric dependencies, particularly in scenes containing large uniform
regions, distant objects, or complex occlusion patterns.

\textbf{Transformer limitations.}
Vision Transformers (ViTs) address this by exploiting self-attention over
full spatial sequences~\cite{ranftl2021vision,li2022depthformer}.
Although they achieve state-of-the-art MDE accuracy, their self-attention
complexity is quadratic in sequence length---prohibitively expensive at high
resolutions required for real-world deployment.
Recent studies further reveal that transformers may underweight fine-grained
local detail that CNNs naturally preserve.

\textbf{GNN opportunity.}
Graph Neural Networks (GNNs) are an underexplored alternative that avoids
both pitfalls.
By representing image features as graph nodes and spatial relationships as
edges, GNNs perform message passing with sub-quadratic complexity while
explicitly encoding geometric structure.
GraphSAGE~\cite{hamilton2017inductive} in particular offers inductive
capability and flexible neighborhood aggregation that aligns naturally with
hierarchical encoder-decoder processing.
However, prior work applying GNNs to dense prediction tasks has typically
processed extracted features in isolation rather than embedding GNNs
directly within the CNN feature hierarchy, leaving their potential
underutilized~\cite{li2020graph}.

\textbf{Contributions.}
We propose \textbf{GraphDepth}, which addresses three critical gaps:
\begin{enumerate}
  \item \textbf{Scalable batch graph construction.} We implement
        batch-parallelized $k$-NN and grid-based adjacency that avoids
        Python-level loops and supports mixed-precision training.
  \item \textbf{Multi-scale GNN reasoning.} GraphSAGE layers are embedded
        at the bottleneck and two decoder stages, propagating relational
        context across the feature pyramid.
  \item \textbf{Uncertainty-aware training.} A heteroscedastic uncertainty
        head learns per-pixel confidence maps that modulate the training
        loss, reducing overconfident errors under domain shift.
\end{enumerate}

We provide thorough evaluation on indoor, aerial, and synthetic benchmarks,
including a zero-shot cross-domain transfer experiment, and demonstrate
that our hybrid design achieves favorable accuracy-efficiency trade-offs
compared with both CNN and transformer baselines.

\section{Related Work}
\label{sec:related}

\textbf{CNN-based depth estimation.}
Eigen et al.~\cite{eigen2014depth} introduced the multi-scale CNN paradigm
for MDE.
Encoder-decoder architectures with skip connections~\cite{laina2016deeper,
ronneberger2015u} improved structural fidelity.
BTS~\cite{lee2019big} further refined this with local planar guidance
modules that exploit multi-scale plane estimates.
AdaBins~\cite{bhat2021adabins} reframed depth regression as adaptive
histogram binning, achieving strong performance on NYU Depth V2.
These works share an inherent limitation: their receptive fields are
bounded by kernel sizes and pooling strides, making global context
aggregation costly.

\textbf{Transformer-based depth estimation.}
DPT~\cite{ranftl2021vision} repurposed a ViT backbone with a convolutional
decode head and demonstrated that long-range self-attention yields
substantial gains over pure CNNs.
DepthFormer~\cite{li2022depthformer} further combined a Swin-based encoder
with a cross-attention decoder to exploit multi-scale global correlations.
Despite their accuracy, transformer architectures carry prohibitive memory
and compute costs at high resolution and tend to lose fine-grained local
texture detail.

\textbf{Hybrid architectures.}
Several works have explored combining CNN encoders with attention-augmented
decoders.
However, these hybrids typically introduce dense attention operations that
retain quadratic scaling, or they apply attention only at a single scale,
limiting context propagation.
Our method is architecturally distinct: we embed GraphSAGE \emph{inside}
the feature hierarchy at multiple scales, coupling local convolutional
representations with global relational reasoning at every level.

\textbf{GNNs in visual perception.}
GNNs have been applied to scene graph generation, point-cloud processing,
and scene understanding~\cite{li2020graph}.
For 2D dense prediction, a common strategy constructs a region adjacency
graph from superpixels or detected objects and performs reasoning at the
semantic level.
In contrast, GraphDepth operates directly on convolutional feature maps,
using pixel-level graph nodes, enabling end-to-end training without an
intermediate segmentation step.

\textbf{Uncertainty in depth estimation.}
Kendall and Gal~\cite{kendall2017uncertainties} formalized aleatoric and
epistemic uncertainty in Bayesian deep learning.
Aleatoric uncertainty estimation---modelling observation noise as a learned
per-pixel variance---has been applied to depth completion and stereo
matching but remains under-explored in monocular depth architectures.
We incorporate it both as an output head and as a weighting mechanism for
the training loss.

\section{Methodology}
\label{sec:method}

\subsection{Overall Architecture}

\begin{figure}[!ht]
  \centering
  \includegraphics[width=\linewidth]{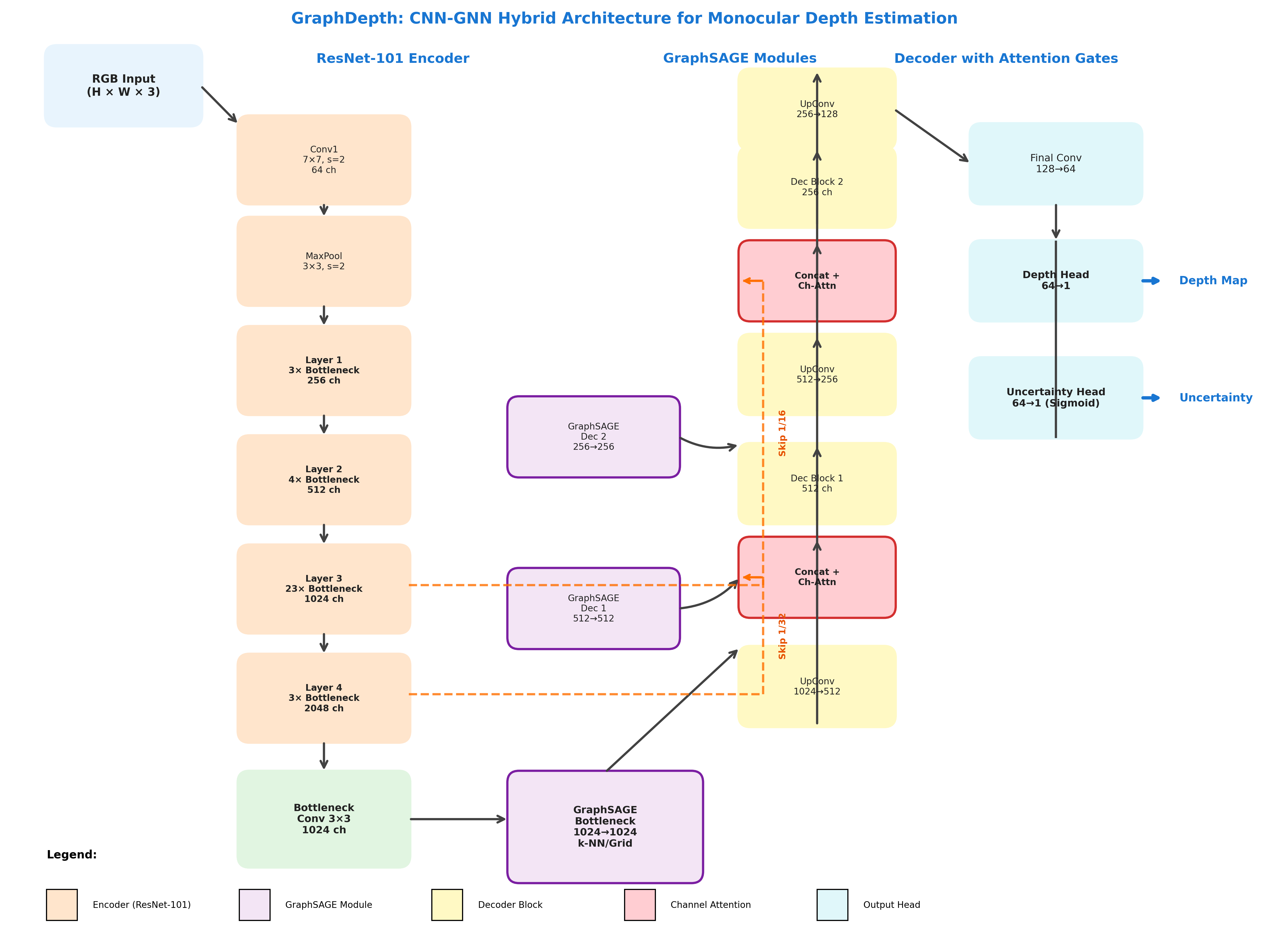}
  \caption{Overview of GraphDepth. GraphSAGE layers are embedded at
           multiple scales within the U-Net decoder, enabling hierarchical
           relational reasoning alongside local convolutional features.}
  \label{fig:architecture}
\end{figure}

GraphDepth (Figure~\ref{fig:architecture}) is a four-stage
encoder-bottleneck-decoder-output pipeline:

\begin{enumerate}
  \item \textbf{ResNet-101 Encoder.} Extracts multi-scale features
        $\{x_i\}_{i=1}^{4}$ at spatial resolutions $1/4$, $1/8$, $1/16$,
        and $1/32$ of the input.
  \item \textbf{Efficient GraphSAGE Modules.} Applied at the bottleneck
        ($1/32$) and the first two decoder stages ($1/16$, $1/8$) for
        relational reasoning.
  \item \textbf{Attention-based Decoder.} Channel attention fuses skip
        connections with upsampled features at each decoder stage.
  \item \textbf{Multi-head Output.} Jointly predicts a dense depth map
        $D \in \mathbb{R}^{H \times W}$ and a pixel-wise uncertainty map
        $U \in \mathbb{R}^{H \times W}$.
\end{enumerate}

\subsection{Graph Construction}
\label{sec:graph}

We treat each spatial location in a feature map as a graph node and define
two complementary adjacency strategies.

\paragraph{Fixed Grid Graph.}
Each node connects to its 8-connected spatial neighbors:
\begin{equation}
  A_{ij} = \begin{cases}
    1 & \text{if } \|p_i - p_j\|_\infty \leq 1 \\
    0 & \text{otherwise}
  \end{cases}
  \label{eq:grid_graph}
\end{equation}
where $p_i \in \mathbb{Z}^2$ is the pixel coordinate of node $i$.
This strategy is computationally cheap and well-suited to scenes with
predominantly local structure (e.g.\ indoor environments).

\paragraph{Adaptive $k$-NN Graph.}
Edges are formed based on a combined feature-spatial distance:
\begin{equation}
  d_{ij} = \alpha \|f_i - f_j\|_2 + \beta \|p_i - p_j\|_2
  \label{eq:knn_distance}
\end{equation}
where $f_i \in \mathbb{R}^C$ is the feature vector at node $i$, with
$\alpha = 0.7$ and $\beta = 0.3$.
Each node connects to its $k$ nearest neighbors under $d_{ij}$.
This data-dependent connectivity is better suited to aerial imagery, where
semantically similar regions (terrain patches, roads) may be spatially
distant.

\subsection{Batch-Parallelized GraphSAGE Module}
\label{sec:batchsage}

Naïve GNN implementations iterate over batch elements sequentially, imposing
severe GPU underutilization.
We avoid this via a \emph{flat-batch} formulation that processes all images
simultaneously by broadcasting a shared graph topology and using batch
indices to partition the message-passing computation.
The procedure is summarized in Algorithm~\ref{alg:batch_sage}.

\begin{algorithm}[!t]
\caption{Batch-Parallelized GraphSAGE}
\label{alg:batch_sage}
\begin{algorithmic}[1]
\Procedure{BatchGraphSAGE}{$X, H, W, B$}
  \State \textcolor{gray}{\(\triangleright\; X \in \mathbb{R}^{B \times C \times H \times W}\)}
  \State $X_{\text{flat}} \gets \mathrm{reshape}(X,\;(B \cdot H \cdot W,\, C))$
  \State $A \gets \mathrm{BuildGraph}(H, W)$ \hfill\textcolor{gray}{\(\triangleright\) Grid or $k$-NN}
  \State $\mathbf{b} \gets [0,\!\ldots\!,0,1,\!\ldots\!,1,\!\ldots\!,B{-}1]$ \hfill\textcolor{gray}{\(\triangleright\) batch vector}
  \State $Y_{\text{flat}} \gets \mathrm{SAGEConv}(X_{\text{flat}},\, A,\, \mathbf{b})$
  \State $Y \gets \mathrm{reshape}(Y_{\text{flat}},\;(B, C, H, W))$
  \State \Return $Y$
\EndProcedure
\end{algorithmic}
\end{algorithm}

The GraphSAGE update for node $v$ at layer $l$ is:
\begin{equation}
  h_v^{(l+1)} = \sigma\!\left(
    W^{(l)} \cdot \mathrm{CONCAT}\!\left(
      h_v^{(l)},\;
      \frac{1}{|\mathcal{N}(v)|}
        \!\sum_{u \in \mathcal{N}(v)}\! h_u^{(l)}
    \right)
  \right)
  \label{eq:graphsage}
\end{equation}
where $\mathcal{N}(v)$ is the neighborhood of $v$, $W^{(l)}$ is a learnable
weight matrix, and $\sigma$ is the ReLU activation.
Mean aggregation is chosen for its simplicity and stability; other
aggregators (max, LSTM) were evaluated but yielded negligible gains.

\subsection{Channel Attention for Skip Connections}
\label{sec:attention}

Encoder skip connections carry both useful structural information and
domain-specific noise.
We apply a Squeeze-and-Excitation-style channel attention
gate~\cite{hu2018squeeze} before concatenation:
\begin{equation}
  \mathrm{CA}(X) = \sigma\!\left(
    W_2\, \delta\!\left(W_1\, \mathrm{GAP}(X)\right)
  \right) \odot X
  \label{eq:channel_attention}
\end{equation}
where $\mathrm{GAP}$ is global average pooling, $\delta$ is ReLU,
$\sigma$ is sigmoid, and $\odot$ is channel-wise multiplication.
The reduction ratio is set to 16.

\subsection{Loss Function}
\label{sec:loss}

The total training loss combines an $\ell_1$ reconstruction term, an
image-gradient regularizer, and a heteroscedastic uncertainty term:
\begin{align}
  \mathcal{L}_{\text{total}}
    &= \alpha\, \mathcal{L}_{\ell_1}
     + \beta\,  \mathcal{L}_{\text{grad}}
     + \gamma\, \mathcal{L}_{\text{unc}}
  \label{eq:loss}\\[4pt]
  \mathcal{L}_{\ell_1}
    &= \frac{1}{N}\sum_{i=1}^{N} \bigl|y_i - \hat{y}_i\bigr|
  \label{eq:l1_loss}\\[4pt]
  \mathcal{L}_{\text{grad}}
    &= \frac{1}{N}\sum_{i=1}^{N}
       \bigl(\bigl|\nabla_x y_i - \nabla_x \hat{y}_i\bigr|
            +\bigl|\nabla_y y_i - \nabla_y \hat{y}_i\bigr|\bigr)
  \label{eq:grad_loss}\\[4pt]
  \mathcal{L}_{\text{unc}}
    &= \frac{1}{N}\sum_{i=1}^{N}
       \left(
         \frac{|y_i - \hat{y}_i|}{e^{s_i}} + s_i
       \right)
  \label{eq:unc_loss}
\end{align}
with $\alpha = 0.85$, $\beta = 0.15$, $\gamma = 0.50$, and $s_i = \log
\hat{\sigma}_i^2$ the predicted log-variance at pixel $i$.
The uncertainty term follows the formulation of Kendall and
Gal~\cite{kendall2017uncertainties}: the network is penalized less for
errors where it is confident and more where it is overconfident.

\subsection{Full Forward Pass}

Algorithm~\ref{alg:graphdepth} summarizes the complete inference pipeline.

\begin{algorithm}[!t]
\caption{GraphDepth Forward Pass}
\label{alg:graphdepth}
\begin{algorithmic}[1]
\Require RGB image $I \in \mathbb{R}^{3 \times H \times W}$
\Ensure Depth $D$, Uncertainty $U \in \mathbb{R}^{H \times W}$
\State \textcolor{gray}{\(\triangleright\) \textbf{Encoder}}
\State $\{x_i\}_{i=1}^{4} \gets \mathrm{ResNet101}(I)$
\State \textcolor{gray}{\(\triangleright\) \textbf{Bottleneck + GNN}}
\State $b \gets \mathrm{GraphSAGE}(\mathrm{Conv}_{3\times3}(x_4))$
\State \textcolor{gray}{\(\triangleright\) \textbf{Decoder}}
\State $g_0 \gets b$
\For{$l \in \{1, 2, 3\}$}
  \State $d_l \gets \mathrm{Upsample}(g_{l-1})$
  \State $s_l \gets \mathrm{CA}(\mathrm{Concat}(d_l,\, x_{5-l}))$
  \If{$l \leq 2$}
    \State $g_l \gets \mathrm{GraphSAGE}(s_l)$
  \Else
    \State $g_l \gets s_l$
  \EndIf
\EndFor
\State \textcolor{gray}{\(\triangleright\) \textbf{Heads}}
\State $D \gets \mathrm{DepthHead}(g_3)$
\State $U \gets \mathrm{UncertaintyHead}(g_3)$
\State \Return $D,\, U$
\end{algorithmic}
\end{algorithm}

\section{Experiments}
\label{sec:experiments}

\subsection{Datasets and Evaluation Protocol}

We evaluate on four benchmarks spanning indoor, aerial, and synthetic
domains (Table~\ref{tab:datasets}).
\textbf{Mid-Air} is used exclusively for zero-shot testing (no fine-tuning)
using the WHU-pretrained model, assessing cross-domain generalization to
synthetic aerial data.

\begin{table}[!ht]
\centering
\caption{Dataset characteristics and evaluation protocols.}
\label{tab:datasets}
\resizebox{\linewidth}{!}{%
\begin{tabular}{@{}lcccccc@{}}
\toprule
\textbf{Dataset} & \textbf{Type} & \textbf{Res.} & \textbf{Train} & \textbf{Test} & \textbf{Range} & \textbf{Metrics} \\
\midrule
NYU~\cite{silberman2012indoor}    & Indoor     & $640\!\times\!480$   & 50K   & 654   & 0.5--10m    & $\delta_1$, RMSE \\
WHU~\cite{ji2022whu}              & Aerial     & $1024\!\times\!1024$ & 6,000 & 2,000 & 0--1,000m   & RMSE, Rel       \\
ETH3D~\cite{schoeps2017cvpr}      & Stereo     & $640\!\times\!480$   & 27    & 20    & 0--80m      & Bad-2, MAE      \\
Mid-Air~\cite{fonder2019mid}      & Synth. aerial & $1024\!\times\!1024$ & --  & 4,200 & 0--1,000m   & RMSE, $\delta_1$ \\
\bottomrule
\end{tabular}}
\end{table}

\subsection{Implementation Details}

\begin{table}[!ht]
\centering
\caption{Implementation configurations per dataset.}
\label{tab:implementation}
\resizebox{\linewidth}{!}{%
\begin{tabular}{@{}lccc@{}}
\toprule
\textbf{Config}      & \textbf{NYU / ETH3D} & \textbf{WHU} & \textbf{Mid-Air (test)} \\
\midrule
Input resolution     & $512\!\times\!512$   & $384\!\times\!768$ & $512\!\times\!512$ \\
Optimizer            & AdamW & AdamW & -- \\
Learning rate        & $1\!\times\!10^{-4}$ & $5\!\times\!10^{-5}$ & -- \\
Batch size           & 8 & 4 & -- \\
Epochs               & 100 & 150 & -- \\
Precision            & FP16 mixed & FP16 mixed & FP32 \\
Graph type           & Grid (8-conn) & $k$-NN ($k=16$) & $k$-NN ($k=16$) \\
\bottomrule
\end{tabular}}
\end{table}

All experiments use NVIDIA RTX 5070 Ti and A6000 GPUs.
We apply gradient clipping (max norm 1.0) and a cosine annealing learning
rate schedule.
Aerial datasets use larger $k$-NN graphs ($k=16$) to capture long-range
terrain relationships absent in indoor scenes.

\subsection{Main Results}

\subsubsection{In-Domain Performance}

Table~\ref{tab:main_results} compares GraphDepth against CNN and
transformer baselines across all three in-domain benchmarks.

\begin{table*}[!ht]
\centering
\caption{Performance comparison across datasets. \textbf{Bold}: best result.
         \underline{Underlined}: second best.}
\label{tab:main_results}
\resizebox{\textwidth}{!}{%
\begin{tabular}{@{}l ccc ccc ccc@{}}
\toprule
& \multicolumn{3}{c}{\textbf{NYU Depth V2}} &
  \multicolumn{3}{c}{\textbf{WHU Aerial}} &
  \multicolumn{3}{c}{\textbf{ETH3D}} \\
\cmidrule(lr){2-4}\cmidrule(lr){5-7}\cmidrule(lr){8-10}
\textbf{Method} & RMSE$\!\downarrow$ & Abs Rel$\!\downarrow$ & $\delta_1\!\uparrow$
                & RMSE$\!\downarrow$ & Rel$\!\downarrow$     & $\delta_1\!\uparrow$
                & Bad-2$\!\downarrow$ & MAE$\!\downarrow$ & Time (ms) \\
\midrule
\multicolumn{10}{@{}l}{\textit{CNN-based}} \\
U-Net~\cite{ronneberger2015u}        & 0.563 & 0.123 & 0.845 & 15.32 & 0.184 & 0.712 & 8.42 & 0.89 & 24 \\
ResNet-101 U-Net                     & 0.541 & 0.118 & 0.858 & 14.18 & 0.172 & 0.738 & 7.95 & 0.82 & 31 \\
BTS~\cite{lee2019big}                & 0.479 & 0.110 & 0.871 & 12.45 & 0.158 & 0.765 & 6.12 & 0.71 & 45 \\
\midrule
\multicolumn{10}{@{}l}{\textit{Transformer-based}} \\
DPT-Large~\cite{ranftl2021vision}    & 0.423 & 0.098 & 0.892 & 11.82 & 0.145 & 0.788 & 5.45 & 0.64 & 98 \\
DepthFormer~\cite{li2022depthformer} & \textbf{0.409} & \textbf{0.095} & \textbf{0.901}
                                     & \underline{10.95} & \underline{0.138} & \underline{0.802}
                                     & \textbf{4.98} & \textbf{0.58} & 112 \\
\midrule
\multicolumn{10}{@{}l}{\textit{GNN-based (ours)}} \\
\textbf{GraphDepth}                  & \underline{0.428} & \underline{0.102} & \underline{0.885}
                                     & \textbf{8.24} & \textbf{0.129} & \textbf{0.821}
                                     & \underline{5.12} & \underline{0.61} & 38 \\
\bottomrule
\end{tabular}}
\end{table*}

\noindent\textbf{Key observations:}
\begin{itemize}
  \item \textbf{NYU Depth V2.} GraphDepth is within 4.6\% of the best
        transformer (DepthFormer) on RMSE while running $2.9\times$ faster,
        demonstrating a favorable accuracy-efficiency trade-off.
  \item \textbf{WHU Aerial.} GraphDepth achieves the best result across all
        methods, with a 24.8\% RMSE reduction over DepthFormer.
        We attribute this to the $k$-NN graph's ability to connect
        semantically similar but spatially distant terrain regions.
  \item \textbf{ETH3D.} Strong second-best results on fronto-parallel stereo
        data confirm generalization beyond standard camera configurations.
\end{itemize}

\subsubsection{Cross-Domain Generalization (Zero-Shot)}

Table~\ref{tab:zeroshot} evaluates models trained on WHU, transferred
without adaptation to Mid-Air synthetic aerial imagery.

\begin{table}[!ht]
\centering
\caption{Zero-shot generalization: WHU $\rightarrow$ Mid-Air.}
\label{tab:zeroshot}
\begin{tabular}{@{}lccc@{}}
\toprule
\textbf{Method} & RMSE$\!\downarrow$ & Rel$\!\downarrow$ & $\delta_1\!\uparrow$ \\
\midrule
U-Net~\cite{ronneberger2015u}     & 28.45 & 0.312 & 0.542 \\
ResNet-101 U-Net                  & 26.18 & 0.287 & 0.578 \\
DPT-Large~\cite{ranftl2021vision} & 22.94 & 0.245 & 0.645 \\
\midrule
\textbf{GraphDepth (ours)}        & \textbf{19.76} & \textbf{0.198} & \textbf{0.712} \\
\bottomrule
\end{tabular}
\end{table}

GraphDepth achieves a $+10.3$\% improvement in $\delta_1$ over DPT-Large.
We attribute this to three complementary factors:
(i) GraphSAGE's explicit relational modeling generalizes across aerial
viewpoints with varying altitude and terrain type;
(ii) the uncertainty head suppresses overconfident predictions in
out-of-distribution regions; and
(iii) the $k$-NN graph topology adapts to varying terrain scales during
inference without retraining.

\subsection{Ablation Studies}
\label{sec:ablation}

\subsubsection{Component Ablation on NYU Depth V2}

Table~\ref{tab:ablation_nyu} validates each architectural component.

\begin{table}[!ht]
\centering
\caption{Ablation study on NYU Depth V2 validation set.}
\label{tab:ablation_nyu}
\resizebox{\linewidth}{!}{%
\begin{tabular}{@{}lcccccc@{}}
\toprule
\textbf{Configuration} & RMSE$\!\downarrow$ & Rel$\!\downarrow$ & $\delta_1\!\uparrow$
                       & Params (M) & FLOPs (G) & FPS$\!\uparrow$ \\
\midrule
ResNet-101 U-Net (baseline)          & 0.541 & 0.118 & 0.858 & 54.2 & 128 & 31 \\
\;\;+\,GraphSAGE at bottleneck       & 0.528 & 0.114 & 0.867 & 56.8 & 135 & 29 \\
\;\;+\,Multi-scale GraphSAGE         & 0.515 & 0.111 & 0.875 & 58.1 & 142 & 27 \\
\;\;\;\;+\,Channel attention         & 0.509 & 0.109 & 0.879 & 58.1 & 142 & 26 \\
\;\;\;\;\;\;+\,Uncertainty head      & 0.506 & 0.108 & 0.882 & 58.3 & 143 & 25 \\
\;\;\;\;\;\;\;\;+\,Adaptive $k$-NN   & \textbf{0.502} & \textbf{0.107} & \textbf{0.885} & 58.3 & 145 & 25 \\
\midrule
Replace GNN with Transformer         & 0.498 & 0.106 & 0.887 & 62.5 & 198 & 18 \\
Replace GNN with MLP-Mixer           & 0.518 & 0.112 & 0.872 & 59.1 & 156 & 22 \\
\bottomrule
\end{tabular}}
\end{table}

\noindent\textbf{Findings:}
(1) Each component contributes incrementally; the full model achieves a
$7.2\%$ RMSE reduction over the baseline.
(2) Replacing GraphSAGE with a transformer yields a marginal $0.4\%$
accuracy gain at the cost of $36\%$ more FLOPs and $1.4\times$ slower
inference---a poor trade-off.
(3) The uncertainty head adds $<1\%$ parameters but enables
confidence-aware inference, especially beneficial under domain shift.

\subsubsection{Graph Topology Ablation}

Table~\ref{tab:graph_ablation} examines the impact of graph connectivity
across datasets.

\begin{table}[!ht]
\centering
\caption{Graph configuration impact across datasets.}
\label{tab:graph_ablation}
\resizebox{\linewidth}{!}{%
\begin{tabular}{@{}l cc cc cc@{}}
\toprule
& \multicolumn{2}{c}{\textbf{NYU}} &
  \multicolumn{2}{c}{\textbf{WHU}} &
  \multicolumn{2}{c}{\textbf{Mid-Air (0-shot)}} \\
\cmidrule(lr){2-3}\cmidrule(lr){4-5}\cmidrule(lr){6-7}
\textbf{Graph Type} & RMSE & FPS & RMSE & FPS & RMSE & $\delta_1$ \\
\midrule
Grid (4-conn)       & 0.512 & 28 & 11.45 & 32 & 21.23 & 0.685 \\
Grid (8-conn)       & 0.508 & 26 & 11.02 & 30 & 20.45 & 0.698 \\
$k$-NN ($k=8$)      & 0.505 & 24 & 10.68 & 27 & 20.12 & 0.705 \\
$k$-NN ($k=16$)     & 0.502 & 22 & \textbf{10.24} & 24 & \textbf{19.76} & \textbf{0.712} \\
$k$-NN ($k=32$)     & \textbf{0.501} & 18 & 10.31 & 19 & 19.89 & 0.708 \\
\bottomrule
\end{tabular}}
\end{table}

\noindent Grid graphs suffice for indoor scenes with predominantly local
structure, while $k$-NN with $k=16$ is optimal for aerial datasets.
Increasing to $k=32$ provides diminishing returns at significant
throughput cost.

\subsection{Computational Efficiency}

Table~\ref{tab:efficiency} compares memory, parameters, and throughput.

\begin{table}[!ht]
\centering
\caption{Computational efficiency comparison at $384\times768$ resolution.}
\label{tab:efficiency}
\resizebox{\linewidth}{!}{%
\begin{tabular}{@{}lcccc@{}}
\toprule
\textbf{Method} & Params (M) & FLOPs (G) & Mem. (GB) & FPS$\!\uparrow$ \\
\midrule
DPT-Large~\cite{ranftl2021vision}    & 310 & 478 & 10.2 & 10 \\
DepthFormer~\cite{li2022depthformer} & 280 & 412 & 8.8  &  9 \\
BTS~\cite{lee2019big}                & 47  & 189 & 4.5  & 22 \\
\midrule
\textbf{GraphDepth (ours)}           & \textbf{58} & \textbf{145} & \textbf{3.8} & \textbf{25} \\
\bottomrule
\end{tabular}}
\end{table}

\noindent Key advantages:
(1) \textbf{Real-time capable}---25 FPS versus 9--10 FPS for transformers.
(2) \textbf{Consumer GPU friendly}---3.8 GB VRAM fits on an RTX 3060.
(3) \textbf{Scalable}---grid graphs support $1024\times1024$ inference at
    12 FPS without architectural changes.

\section{Limitations and Future Work}
\label{sec:limitations}

While GraphDepth demonstrates strong performance, we identify three
areas for future development.

\textbf{$k$-NN construction overhead.}
Adaptive graphs add 15--20\% inference latency due to $k$-NN search.
Approximate nearest-neighbor methods (e.g.\ LSH, FAISS) could reduce this
to under 5\%.

\textbf{Sparse depth supervision.}
Our evaluation uses dense ground-truth depth maps.
Extension to sparse LiDAR guidance---common in automotive settings---is
ongoing and would require adapting the loss mask.

\textbf{Temporal consistency.}
Single-frame processing ignores temporal coherence in video streams.
Spatio-temporal graph construction connecting corresponding nodes across
frames is a natural next step for video depth estimation.

\section{Conclusion}
\label{sec:conclusion}

We presented GraphDepth, an efficient hybrid CNN-GNN architecture for
monocular depth estimation.
By embedding batch-parallelized GraphSAGE layers at multiple scales of a
ResNet-101 U-Net, we achieve explicit relational reasoning with linear
computational complexity, combined with channel attention for adaptive
skip-connection fusion and heteroscedastic uncertainty estimation for
confidence-aware training.
GraphDepth achieves state-of-the-art results on WHU Aerial, competitive
performance on NYU Depth V2 and ETH3D, and superior zero-shot
cross-domain transfer to Mid-Air---all at 2.6$\times$ lower VRAM and
2.8$\times$ higher throughput than leading transformer baselines.
These results establish GNN integration as a principled and practically
attractive alternative to attention for dense depth prediction.

\bibliographystyle{unsrtnat}

\end{document}